\documentclass[10pt,twocolumn,letterpaper]{article}

\usepackage{cvpr}              

\usepackage{graphicx}
\usepackage{amsmath}
\usepackage{amssymb}
\usepackage{booktabs}
\usepackage{algorithm}
\usepackage{algpseudocode}

\usepackage[pagebackref,breaklinks,colorlinks]{hyperref}

\usepackage[capitalize]{cleveref}
\crefname{section}{Sec.}{Secs.}
\Crefname{section}{Section}{Sections}
\Crefname{table}{Table}{Tables}
\crefname{table}{Tab.}{Tabs.}

\def\cvprPaperID{33} 
\def\confName{CVPR}
\def\confYear{2022}

\begin{document}

\title{Video-ReTime: Learning Temporally Varying Speediness for Time Remapping}

\author{Simon Jenni \qquad Markus Woodson \qquad Fabian Caba Heilbron\\
Adobe Research\\
{\tt\small \{jenni,woodson,caba\}@adobe.com}
}
\maketitle

\begin{abstract}
We propose a method for generating a temporally remapped video that matches the desired target duration while maximally preserving natural video dynamics.
Our approach trains a neural network through self-supervision to recognize and accurately localize temporally varying changes in the video playback speed. 
To re-time videos, we 1. use the model to infer the slowness of individual video frames, and 2. optimize the temporal frame sub-sampling to be consistent with the model's slowness predictions.  
We demonstrate that this model can detect playback speed variations more accurately while also being orders of magnitude more efficient than prior approaches.
Furthermore, we propose an optimization for video re-timing that enables precise control over the target duration and performs more robustly on longer videos than prior methods.
We evaluate the model quantitatively on artificially speed-up videos, through transfer to action recognition, and qualitatively through user studies.
\end{abstract}

\section{Introduction}
\label{sec:intro}

Satisfying requirements on the duration of videos is a common problem for video content producers: customers might request a fixed video length, social media platforms have strict limits on video uploads, or video transitions should align with music transitions, to name just a few.
Fulfilling these constraints through video editing is time-consuming and requires considerable skills. 
Video content consumers likewise often desire to save time and consume videos in a shorter time, as indicated by the presence of speed controls in popular video platforms. 
The straightforward solution to this problem is to speed up the video uniformly to achieve the desired target duration instead of editing it. 
While this naive approach works fine in video segments without much motion, the result might look hectic and unrealistic in more action-packed scenes.

However, many video scenes exhibit a range of playback speeds that can be considered natural. 
While some scenes are restricted due to the laws of physics (\eg, a person running or jumping), other scenes do not have such tight constraints (\eg, a car driving at constant speed or a static scene with a moving camera). 
This suggests that a non-uniform speed-up of videos could often provide a natural-looking result. 
Detecting such slow scenes requires a good understanding of the scene content and their natural movement patterns and can not be based solely on motion magnitude, \eg, via optical flow.
Furthermore, the parts of a video showing such ambiguous speediness are often temporally tightly localized, making manual detection and speeding up of such regions intractable.

\begin{figure*}[t]
    \centering
    \includegraphics[width=0.78\linewidth]{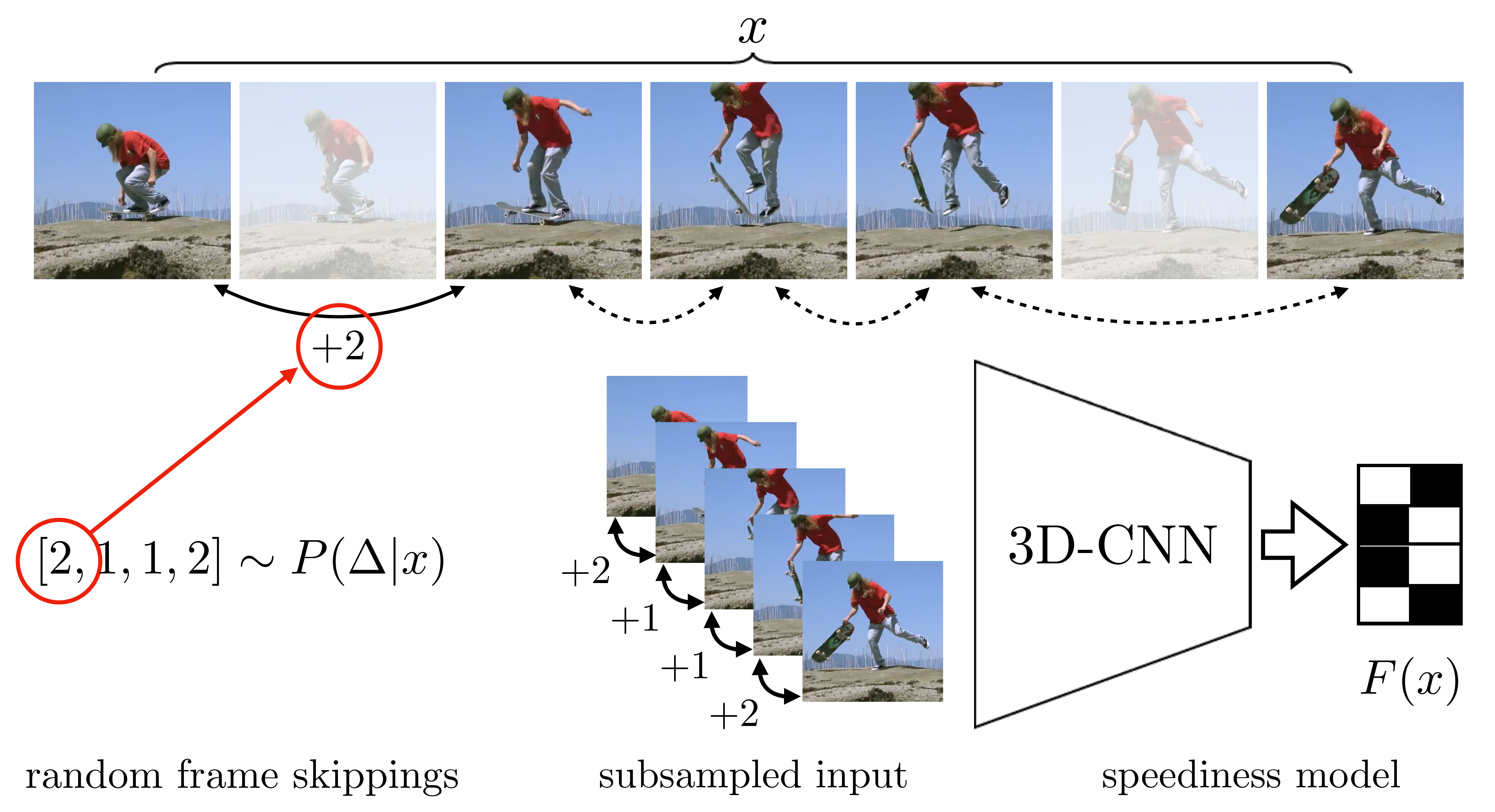}
    \caption{\textbf{Self-supervised time-varying speediness prediction.} 
    We train a neural network $F$ to recognize time-varying speed changes in videos. 
    Given a video $x$, we sample a sequence of random frame skippings, sub-sample a training video accordingly, and train $F$ to classify the frame skips in a self-supervised fashion.
    Note that our model makes a speed prediction between every pair of consecutive frames and thus preserves the temporal dimension of the input.
    }
    \label{fig:ssl}
\end{figure*}

In this work, we describe a method to automatically find such temporally local video segments that can be sped up while preserving realism in the video dynamics. 
Our method relies on a neural network trained via self-supervision to recognize and localize artificially introduced changes in the playback speed of video frames. 
In fact, prior works have shown that self-supervised playback speed recognition learns good video representations \cite{epstein2020oops,benaim2020speednet,yao2020video,jenni2020video} and that these speed predictions can be leveraged for re-timing \cite{benaim2020speednet}. 
In contrast to these works, we train the model on temporally \emph{non-uniform} frame speeds and infer the speed of each frame in the input sequence rather than a single speed common to all frames in the sequence (see Figure~\ref{fig:ssl}). 
To this end, we modify existing 3D-CNN architectures to adapt them to the dense speed prediction task and enable accurate per-frame predictions.  
These modifications make our approach both more accurate in localizing speed changes (\cite{benaim2020speednet} assigns predictions to the middle frame of the input frame sequence) and more efficient (our method does not require a sliding window approach). 

Given the pre-trained speediness network, the process of re-timing videos has two stages (see Figure \ref{fig:retime}). 
First, we compute slowness predictions to identify frames that can be sped up (this has to be done once per video).
The second stage then optimizes the video sub-sampling to achieve the target duration based on the slowness predictions (done once per target duration). 
For the optimization in the second stage, we propose an objective function that directly optimizes the sub-sampling given a chosen target duration.  
This is in contrast to the method proposed in \cite{benaim2020speednet}, where the optimization is based on achieving a target per-frame speed-up factor averaged over the source video sequence. 
Only after the optimization is this per-frame speed-up converted into a subsampling. 
We observed two shortcomings of this approach: 1. it provides less accurate control over the output video duration (the duration of the sub-sampled video depends on the temporal distribution of the per-frame speed-ups), and 2. error accumulation in the frame speed-ups, which leads to degrading performance, especially with longer videos.
Our solution instead verifies that the sub-sampling is consistent with the speediness predictions throughout the optimization and thus performs more accurately and robustly when applied to longer video clips. 
Furthermore, we also describe how to modify this objective to work with any re-timing signal (\eg, using frame feature similarity, which we explore in experiments) and how to control the strength of the effect. 

In our experiments, we demonstrate that our model is better able to recognize time-varying speed-ups on unseen videos of the Kinetics \cite{zisserman2017kinetics} dataset and also learns better features than models trained on uniformly speed-up videos.
We also demonstrate improved re-timing performance given synthetic ground-truth speed predictions, especially on longer videos. 
Finally, we report the results of user studies where our re-timing was judged more natural than a naive uniform re-sampling and was also preferred to a re-timing based on frame similarity. \\

\noindent \textbf{Contributions.} To summarize, we make the following contributions: 1) We propose a model architecture and self-supervised learning task to learn time-varying speediness from unlabelled videos; 2) We propose an objective function that takes speediness or any other re-timing signal and computes a frame subsampling that satisfies a user-defined target duration; 3) We demonstrate in experiments that our approach is more accurate at localizing speed changes, scales better to longer input videos, and is more computationally efficient than prior approaches.

\section{Prior Work}

\noindent \textbf{Self-Supervised Learning.}
Our model leverages self-supervision to automatically build data for the speediness model training.
Learning tasks that do not require human labor for supervision have been popularized as pretext tasks for image representation learning.
Classic examples are the ordering of image patches \cite{Carl2015,noroozi2016unsupervised}, the colorization of gray-scale images \cite{zhang2016colorful,zhang2016split}, or the recognition of sets of image transformations \cite{gidaris2018unsupervised,jenni2020steering}.
Similar pretext tasks have also been proposed on video, \eg, ordering of video frames \cite{misra2016shuffle,brattoli2017lstm,fernando2017self,lee2017unsupervised}, recognizing the playback direction \cite{wei2018learning}.
Most relevant to this work, classifying the playback speeds of videos \cite{epstein2020oops,benaim2020speednet,yao2020video,jenni2020video} was shown to be a useful pretext task.
More recently, such temporal learning tasks have been combined with contrastive learning \cite{chen2020simple,he2020momentum} to boost the performance of the learned video representations \cite{dave2021tclr,patrick2020multi,bai2020can,jenni2021time}.
In contrast to those prior works, we consider non-uniform speed-ups of the videos and propose a dense per-frame prediction task. 
Notably, \cite{jenni2020video} also considers non-uniform temporal warpings of videos but instead learns to classify examples of this type as a separate category distinct from other speed classes. 
Furthermore, our focus is on the application for video re-timing rather than general video representation learning. 
Indeed, we find that while combining speed prediction with contrastive objectives boosts feature performance in transfer to action recognition, we observe a degradation of the speed prediction performance. \\

\noindent \textbf{Temporal Video Remapping.}
Time remapping has some relation to the rich body of work on video summarization \cite{apostolidis2021video,smith1997video}.
Typically, summarization aims at selecting only a few keyframes or video fragments that best represent the content of the video and thus potentially discard a large chunk of the video.
Our goal is instead to only remove small temporal fragments (typically only a few consecutive frames) in such a manner that the result still preserves natural motion as much as possible.
Such time remapping has been explored before, \eg, \cite{bennett2007computational} use various visual metrics between pairs of frames to sub-sample a video while satisfying a chosen target duration.
Other works focused on fast-forwarding of egocentric videos \cite{poleg2015egosampling}, fast-forwarding based on user-chosen query clips \cite{petrovic2005adaptive}, or the case of high frame-rate input videos \cite{zhou2014time}.
The most related prior work is \cite{benaim2020speednet}, which also leverages a learned speediness model for time remapping.
Aside from differences in the sub-sampling optimization, our approach is different since we learn from non-uniformly speed-up videos and output a sequence of dense per-frame speediness predictions. 
These changes enable more accurate localization of speed changes and make our approach more computationally efficient.

\section{Model}

Our goal is to train a neural network $F$ to predict the playback speed between consecutive frames of a video and then leverage the predictions of this network to remap video clips temporally. 
We first describe the self-supervised training of $F$ in Section~\ref{sec:ssl} and outline the optimization of the frame sub-sampling in Section~\ref{sec:retime}.

\subsection{Learning Temporally Varying Speediness}\label{sec:ssl}


Let $\mathbf{x}_i=[x_1, \ldots, x_{n_i}]$ denote a training video consisting of $n_i$ frames. 
To train our model we sample a sequence of frame skips ${\Delta}=[\delta_1, \ldots, \delta_m]$ according to a random procedure $\Delta \sim P(\Delta|\mathbf{x}_i)$.
In our implementation we choose $\delta_i \in \{2^j\}_{j=0}^k$, where $k$ defines the number of speedup classes (we set $k=2$ in experiments). 
To produce a sample of $\Delta$, we first sample $\sigma \sim \mathcal{U}([0, 0.1])$ defining the probability of observing a change in playback speed. We then model $P(\Delta|\mathbf{x}_i, \sigma)$ as a Markov chain with transition probabilities given by
\begin{equation*}
    P(\delta_{i+1}|\delta_{i}, \sigma) = \begin{cases} 
    1-\sigma &\text{if } \delta_{i+1}=\delta_{i}\\ 
    \frac{\sigma}{k} &\text{otherwise}
    \end{cases}.
\end{equation*}

The sequence of frame skips $\Delta$ defines the temporally varying speedups and provides the target classes during self-supervised training via $\mathbf{y}=[\log_2(\delta_1), \ldots, \log_2(\delta_m)]$.     
The frame skips are then translated into frame indices via $\mathbf{\nu}(\mathbf{x}_i, \Delta)=\rho + [0, \delta_1, \ldots, \sum_{i=1}^m \delta_m ]$, where $\rho \sim \mathcal{U} ([0, n_i - \sum_{i=1}^m \delta_i])$ is a random offset. 
Finally, the temporally sub-sampled video clip is given by $\mathbf{x}_i^\nu=[x_{\nu_0}, \ldots, x_{\nu_m}]$, where the superscript $\nu$ indicates the used sub-sampling.  

The model $F$ is constructed such that it preserves the temporal dimension of the input clips, \ie, for an input $\mathbf{x}^\nu \in \mathbb{R}^{(m+1) \times h \times w \times c}$ the output of the network will be $F(\mathbf{x}^\nu) \in \mathbb{R}^{m \times k}$.

We apply a \texttt{softmax} activation to the network output and will thus interpret it as $F(\mathbf{x}^\nu)[i,j]=p(y_i=j|\mathbf{x}^\nu)$ (recall that $\delta_i=2^{y_i}$). 
The training objective is given by a weighted cross-entropy classification loss
\begin{equation}
     \mathop{\mathbb{E}}_{\mathbf{x}_i^\nu, \mathbf{y}_i} \Big[ -\sum_{j=1}^m \omega_{y_j} \cdot \log F(\mathbf{x}_i^\nu)[j,y_j] \Big],
\end{equation}
where the weights $\omega_{y}$ are chosen such that lower speed classes have a higher weight.  
The purpose of this is to bias the model towards predicting lower playback speeds in ambiguous cases and where plausible. 
We set $\omega_{y}=1.25^{(k-y)}$ in our experiments. 
The self-supervised pre-training is illustrated in Figure~\ref{fig:ssl}.

\begin{figure}[t]
    \centering
    \includegraphics[width=\linewidth]{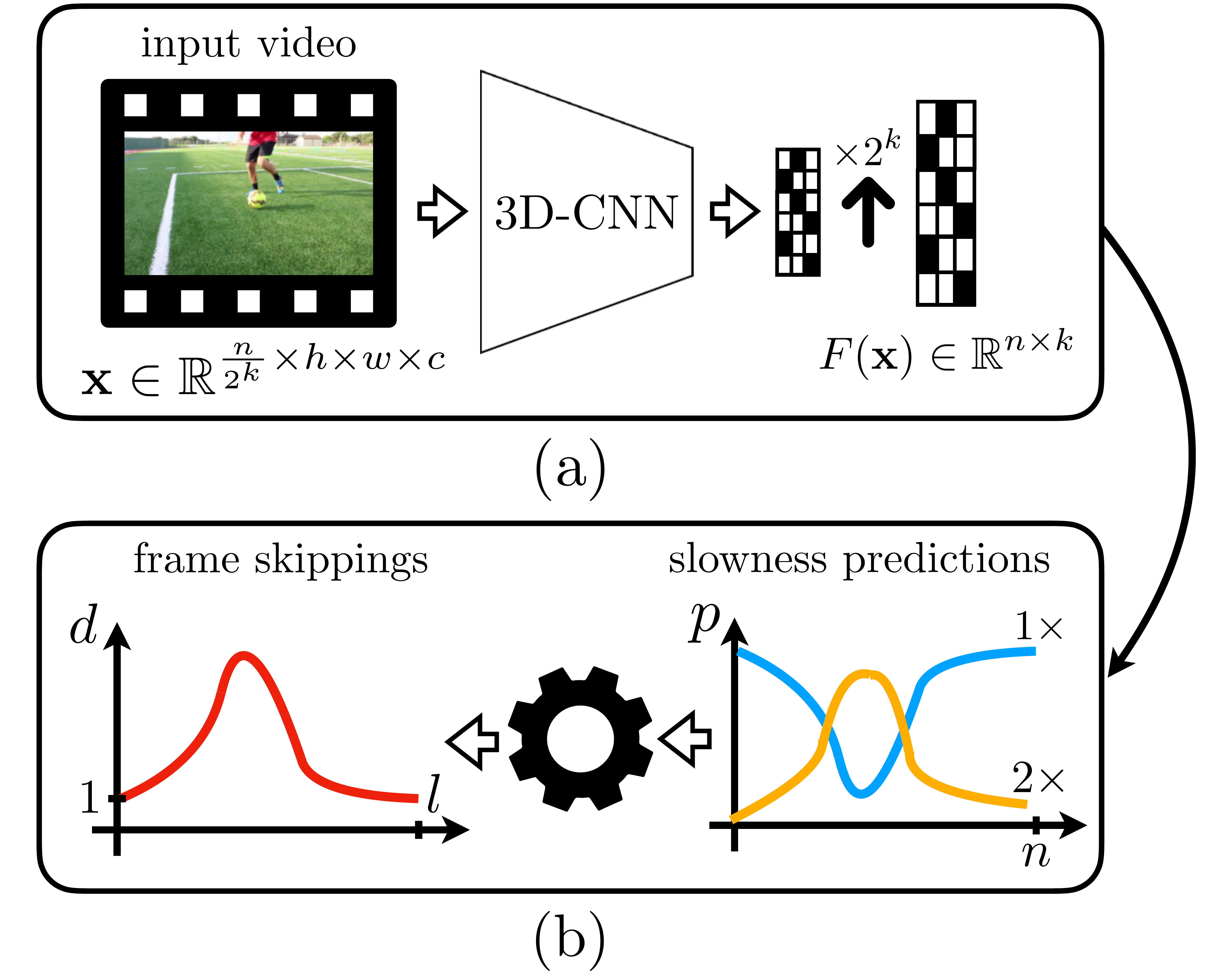}
    \caption{\textbf{Procedure for video re-timing.} 
    Our approach to re-time videos works in two stages: (a) predict the frame slowness of videos played at $2^k$ the original speed ($k=2$ usually). These predictions are then linearly interpolated to the source video length. This step has to be done once per video. (b) Given the slowness likelihoods, optimize the frame sub-sampling to match a user-defined target duration $l$. 
    }
    \label{fig:retime}
\end{figure}

\subsection{Re-Timing Videos} \label{sec:retime}
We want to use the trained speediness model $F$ to re-sample video frames such that the result has the desired target duration and the frame subsampling looks as plausible as possible. 
Given a video $\mathbf{x}=[x_1, \ldots, x_{n}]$ and target length $l<n$, we now describe how to compute between-frame speed likelihoods.
Since we want to detect frames that appear slow, we feed the video at maximum speed $2^k$ through the network. 
This is equivalent to prepending a temporal subsampling layer with stride $2^k$ to the network $F$. 
Let $F_{2^k}$ denote the prepended network, then we have $F_{2^k}(\mathbf{x})\in \mathbb{R}^{\frac{n}{2^k}\times k}$ which we then linearly interpolate to length $n-1$ to get the between-frame speed likelihoods $\mathbf{p}\in \mathbb{R}^{(n-1)\times k}$. 
Note that $\mathbf{p}[i, j] = p(\delta_i=\frac{1}{2^{k-j}})$, \ie, $\mathbf{p}$ provides the likelihood of "slowness" per frame.

Given these slowness predictions, we now describe how to optimize the frame subsampling.
Our goal is to find a sequence of frame skippings $\mathbf{d}=[d_1, \ldots, d_l]$ which then translate into the subsampling indices $\nu=[0, d_1, \ldots, \sum_{i=1}^l d_l ]$ used for video re-timing. 
The objective of $\mathbf{d}$ is composed of four terms
\begin{equation}
    \label{eq:opt}
    \mathcal{L}_d = \mathcal{L}_p  + \lambda_{sum} \mathcal{L}_{sum} + \lambda_{min} \mathcal{L}_{min} + \lambda_{smooth} \mathcal{L}_{smooth}.  
\end{equation}
The term $\mathcal{L}_{sum}$ enforces that the subsampled sequence covers the whole video and is given by
\begin{equation}
   \mathcal{L}_{sum} = \Big|\sum_{i=1}^l d_i - n \Big|^2.
\end{equation}
The term $\mathcal{L}_{min}$ ensures a that frames are not slowed down, \ie, at minimum the original speed is preserved 
\begin{equation}
   \mathcal{L}_{min} = \sum_{i=1}^l \max(0, 1-d_i).
\end{equation}
To control the smoothness of the speed changes, we also introduce a smoothness term 
\begin{equation}
    \mathcal{L}_{smooth} = \sum_{i=1}^{l-1} |d_{i+1}-d_i|^2,
\end{equation}
where $ \lambda_{smooth}$ is a user-chosen parameter. 
Finally, the speediness term $\mathcal{L}_p$ penalizes speedups that go beyond the estimated per-frame slowness predictions and is given by
\begin{equation}\label{eq:target}
    \mathcal{L}_p = \sum_{i=1}^{l-1} \sum_{j=0}^{k}\mathbf{p}[\nu_i, j]\max(0, d_i-2^{k-j})^2.
\end{equation}
Note that the speediness predictions $\mathbf{p}$ are indexed with the estimated subsampling $\nu$. 
This is crucial and ensures that the sub-sampling is consistent with the slowness predictions in the original video.
Since $\nu \in \mathbb{R}^l$ we compute the value of $\mathbf{p}[\nu_i, j]$ through linear interpolation.  

The optimization of Eq.~\ref{eq:opt} starts from an initial uniform sub-sampling $\mathbf{d}_{init}=[\frac{n}{l}, \ldots, \frac{n}{l}]$ and is performed via gradient descent using the Adam optimizer.
Since the loss in Eq.~\ref{eq:target} is sensitive to the confidence of $\mathbf{p}$, \ie, the range of the predicted probabilities, we lower the softmax temperature (hence increase the confidence) such that $\max(\mathbf{p})-\min(\mathbf{p})>0.975$ to achieve consistent behaviour.
The re-timing procedure is illustrated in Figure~\ref{fig:retime}.

\subsection{General Re-Timing Signals}\label{sec:general}

Frequently, the desired duration reduction is more than what can be satisfied by the constraint in Eq.~\ref{eq:target}, \ie, the speed model does not detect enough slow parts or the detected regions are not "slow" enough.
In this case, the model will add an additional speedup uniformly to all the frames.
This might not always be desired. For example, a user might prefer to keep the fastest regions at their original speed while potentially exaggerating the speedup effect in slower parts.

Furthermore, we note that video content editors might have different intentions than preserving natural motion when re-timing a video.
For example, users might want to keep selected segments at their unaltered original speed while they are indifferent about others. 
Another desired effect could be limiting the frame-to-frame dissimilarity, \ie, not speeding up when there are significant changes in scene content due to hectic passages or large camera motion.

To enable more flexibility in the re-timing, we propose an alternative to the loss in Eq.~\ref{eq:target}, where any temporal signal $\mathbf{s}_i=[s_1, \ldots, s_{n_i}]$ (corresponding to the video frames) can be leveraged for re-timing. 
Let us first denote with $\mathbf{\hat{s}}_i$ the signal normalized to the range $[0, 1]$, \ie,
\begin{equation}\label{eq:signal}
    \mathbf{\hat{s}}_i = \frac{\mathbf{s}_i-\min(\mathbf{s}_i)}{\max(\mathbf{s}_i)-\min(\mathbf{s}_i)}.
\end{equation}
Let $\hat{s}_i=0$ indicate a frame that should not be speed-up (otherwise consider $1-\mathbf{\hat{s}}_i$ as the signal).
We then define the loss associated with the general re-timing signal $\mathbf{s}_i$ as,
\begin{equation}
    \mathcal{L}_s = \sum_{i=1}^{l-1} \max(0, d_i-\lambda \mathbf{\hat{s}}[\nu_i] - 1)^2,
\end{equation}
where $\lambda$ is a multiplier that controls the strength of the effect. 
By default we choose $\lambda=\frac{n}{l}\frac{1}{\Bar{\mathbf{s}}}$, where $\Bar{\mathbf{s}}$ is the mean of $\hat{\mathbf{s}}$.
We thus increase the strength of the effect inversely proportional to the desired duration reduction rate $\frac{n}{l}$.

Instead of leveraging the predicted per-frame speediness predictions directly as in Eq.~\ref{eq:target}, we can first define $\mathbf{s}_i=\sum_{j=0}^{k}\mathbf{p}[i, j]2^{k-j}$ as the re-timing signal and then optimize Eq.~\ref{eq:signal}.
This then allows us to control the strength of the effect through an adjustment of $\lambda$.
Likewise, video segments can be kept at their original speed by setting $s_i=0$ for user-selected frames. 
Alternatively, we can leverage other signals, \eg, per-frame classifier scores or frame-to-frame dissimilarity in some feature space.
In the latter case we would, for example, set
\begin{equation}
    \mathbf{s}_i=\frac{\phi(x_i)^T\phi(x_{i+1})}{||\phi(x_i)||\cdot||\phi(x_i)||} ,
\end{equation}
and speed up frames more if their image features $\phi(x_i)$ have a high cosine similarity with the following frames.

\subsection{Model Architecture}\label{sec:tempdiff}
We base the network architecture of $F$ on established 3D-CNN backbone architectures \cite{tran2018closer}.
To preserve the temporal dimension of the input, we set all the temporal strides to $1$ and use 'SAME'-padding throughout the network. 
We further introduce custom network heads tailored to our between-frame speed prediction task. 
The global spatio-temporal average pooling is replaced with spatial-only global average pooling. 
This is followed by two layers of 1D-convolutions.
Since the speed prediction task heavily depends on the temporal difference of the frame contents, we further propose to add temporal differences as an additional input to the model.
Concretely, let $\mathbf{a}_{0:t}^l \in \mathbb{R}^{(t+1)\times h \times w\times c}$ denote the activations at layer $l$ of the network.
We propose to augment the activations at some layer by concatenting their temporal difference before feeding it to the next layer, \ie, we set the activations to $\mathbf{\hat{a}}^l= [\mathbf{a}_{0:t-1}^l, \mathbf{a}_{1:t}^l-\mathbf{a}_{0:t-1}^l]$, where $[\cdot,\cdot]$ is a concatenation along the channel dimension.
By default, we apply this operation before the spatial pooling, but we explore other options in ablations (see Table~\ref{tab:ablations}).

\subsection{Implementation Details}
Networks were trained using the AdamW optimizer \cite{loshchilov2018decoupled} with default parameters and a weight decay of $10^{-4}$. 
During pre-training, the learning rate is first linearly increased to $3\cdot10^{-4}$ and then decayed with a cosine annealing schedule \cite{loshchilov2016sgdr}.
We use standard data augmentations, \eg, horizontal flipping, color-jittering, and random cropping. 
As backbone architecture we use R(2+1)D-18 \cite{tran2018closer}.
Input clips contain  $m=32$ frames at a resolution of $112\times 112$.  
The speediness prediction network was trained on Kinetics \cite{zisserman2017kinetics}.
We set $k=2$ and thus predict frame skippings in $\{1, 2, 4\}$.

\section{Experiments}

We performed experiments to verify the effectiveness of both our proposed temporally varying frame speediness prediction and the subsequent optimization for video re-timing based on inferred slowness likelihoods.
The two stages of our pipeline are evaluated separately in controlled experiments. 
In Section~\ref{sec:expspeed} we evaluate the speediness prediction accuracy and compare it to the sliding window approach of prior work.
Section~\ref{sec:expretime} then investigates the effectiveness of our sub-sampling optimization and compares it to prior work and the naive uniform speed-up.
In Section~\ref{sec:abl} we explore the influence of different self-supervised model configurations in terms of speed prediction and transfer learning performance.
Finally, we report the results of user studies in Section~\ref{sec:user}.

\begin{table}[t]
    \centering
    \caption{\textbf{Time-varying speediness prediction accuracy.} 
    We compare our dense time-varying speediness prediction approach to a sliding window approach proposed in prior work \cite{benaim2020speednet}. In addition, we report the mean classification accuracy over artificially sped-up test videos of Kinetics. We also report relative inference time comparison, showing that, on average, our approach is around 12 times as fast as a sliding window approach. 
    }

    \begin{tabular}{@{}l@{\hspace{2em}}c@{\hspace{1em}}c@{\hspace{1em}}c@{}}
    \toprule
      &   \multicolumn{2}{c}{\textbf{Test Dataset}} &   \\ 
    \textbf{Method}    & \textbf{Kinetics-U}  &   \textbf{Kinetics-V} &   \textbf{Run-time} \\
    \midrule 
    Sliding  ($w=8$)  &   73.6  & 61.9  \\  
    Sliding  ($w=16$)  &  \underline{82.8}   & 62.9 & $12\times$ \\ 
    Sliding  ($w=32$)  &  \textbf{87.0}  & 59.6  \\  

    \midrule
    \textbf{Ours} ($w=8$)  &  79.6   &  83.6  \\
    \textbf{Ours} ($w=16$)  &  81.2   &  \underline{85.6} & $1\times$  \\ 
    \textbf{Ours} ($w=32$)  &  82.6   &  \textbf{87.4}  \\

    \bottomrule
    \end{tabular}
    \label{tab:speed}
\end{table}

\subsection{Speediness Prediction Performance}\label{sec:expspeed}
In order to evaluate the ability of our model to classify and localize time-varying speed changes in videos accurately, we artificially re-sample unseen test videos of Kinetics \cite{zisserman2017kinetics} similarly to our self-supervised pre-training. 
The test data thus consists of videos $\mathbf{x}_i=[x_{0}, x_{\nu_1}, \ldots, x_{\nu_m}]$ where $\nu_j=\sum_{i=0}^j 2^{y_i}$ and the model has to accurately predict the speed classes $y_i$.
The models were pre-trained for 60 epochs on Kinetics.

Table~\ref{tab:speed} shows the results of our proposed model and compares it to models using the sliding window approach used in prior work. 
The sliding window models are based on the same network architecture but were trained on uniformly sped-up video clips (using the same speed classes as our method).
We also compare on such uniformly sped-up videos and refer to the version of Kinetics with varying playback speeds as Kinetics-V and the one with uniform speed-up as Kinetics-U. 
We report results for different temporal windows (\ie, number of input frames) for both models. 

We observe a trade-off of accuracy and localization ability in the sliding window approach: larger windows increase accuracy on Kinetics-U but hurt in the case of Kinetics-V, where localization is essential. 
In contrast, our method benefits from the extended temporal receptive field in both cases and significantly outperforms on Kinetics-V.
We also report the relative run-time improvements achieved through our approach.

\begin{table}[t]
    \centering
    \caption{\textbf{Video re-timing performance.} We compare our temporal sub-sampling optimization to the approach proposed in \cite{benaim2020speednet} on synthetic ground-truth speediness predictions and corresponding target frame skip sequences. We report the mean absolute error of the predicted sequence of frame skips for different sequence lengths.    }
    \begin{tabular}{@{}l@{\hspace{2em}}c@{\hspace{1em}}c@{\hspace{1em}}c@{}}
    \toprule
      &   \multicolumn{3}{c}{\textbf{Clip Duration}}   \\ 
    \textbf{Method}    & \textbf{20sec}  &   \textbf{1min} &   \textbf{3min} \\
    \midrule 
    Uniform speed-up     &  0.934   & 0.907  & 0.917 \\ 
    SpeedNet \cite{benaim2020speednet}   &  \underline{0.229}   & \underline{0.366}  & \underline{0.439} \\ 
    \midrule 
    \textbf{Ours}   &  \textbf{0.156}   &  \textbf{0.227} & \textbf{0.239} \\
    \bottomrule
    \end{tabular}
    \label{tab:retime}
\end{table}

\subsection{Temporal Remapping Accuracy}\label{sec:expretime}
To evaluate the accuracy of the re-timing optimization, we sample sets of ground truth target frame skips $\mathbf{d}\in \mathbb{N}^l$ and produce the corresponding slowness predictions $\mathbf{p} \in \mathbb{R}^{m\times k}$, where $m=\sum_i d_i$ and $\mathbf{p}$ corresponds to a perfect prediction of the frame speed.
We then optimize for the best estimate $\mathbf{\hat{d}}$ using our proposed objective in Eq.~\ref{eq:opt} and report the mean absolute error of the prediction in Table~\ref{tab:retime}. 
We also report the error for a naive uniform speed-up and compare it to an optimization of the per-frame speed-up similar to the approach in \cite{benaim2020speednet}. 

As mentioned before, optimizing for a target per-frame speed-up averaged over the $m$ frames of the input sequence does not directly achieve the desired target duration.
Since the relation between average per-frame speed-up and target duration is non-trivial (it depends on the distribution of the speed-ups over time), we run the optimization of \cite{benaim2020speednet} multiple times with a target speed-up of $(m/l)*1.05^k$ for $k=0, \ldots, 10$ and report the best result. 
We observe that our method is both more accurate and more robust when applied to longer video clips 

\subsection{Ablation Experiments}\label{sec:abl}

We perform ablation experiments to investigate the influence of the time-varying speed prediction task and the proposed architectural design on both downstream transfer learning performance and speed prediction accuracy. 
Models are pre-trained on Kinetics for 30 epochs in these experiments.
To measure the quality of the learned video representations, we transfer the representation to action recognition and retrieval on UCF101 \cite{soomro2012ucf101} and HMDB51 \cite{hmdb51}.
We believe that good transfer performance to action recognition might be a good indicator that the model learned the natural speed of objects instead of relying on low-level cues or artifacts present in the input (\eg, camera motion).
We report results for the following set of ablation experiments in Table~\ref{tab:ablations}:\\

\noindent \textbf{(a)-(c) Pre-training signals:} 
We compare a model that is trained using only uniformly speed-up videos (a) to a model that is trained to predict time-varying speed-ups (b) and a model that additionally performs contrastive learning \cite{chen2020simple} (c). 
We observe that time-varying speed prediction leads both to better features (as measured by action recognition transfer) and better speed prediction accuracy on Kinetics-V.
We also observe that a multitask approach, where speed prediction is combined with contrastive learning, improves feature performance but significantly hurts speed prediction accuracy. 
While prior work has shown that a combination of contrastive learning with speed predictions and other temporal pretext tasks improves action recognition performance \cite{jenni2021time}, our downstream application for re-timing does not seem to benefit from such a multitask pre-training. \\

\noindent \textbf{(d)-(f) Addition of temporal differences:}
Here we investigate the influence of augmenting network activations at different layers with temporal differences as described in Section~\ref{sec:tempdiff}.
We compare a model without temporal differences (d) to a model with temporal differences at the input (e) and one where it is applied in the feature space (f) before the spatial pooling operation. 
We observe that adding temporal differences to the input severely hurts feature performance in transfer while actually performing best on the speed-prediction task. 
Interestingly, adding temporal differences in the feature space improves performance in transfer while preserving speed-prediction accuracy. 
We believe that adding temporal differences to the input increases the network's sensitivity to low-level cues, such as camera motion or encoding artifacts. In contrast, temporal differences in the feature space encourage the learning of input dynamics patterns at a higher level.
Indeed, we observe that the re-timing results of (f) are qualitatively superior to those achieved by (e) most of the time, as they tend to focus on the movement of (foreground) objects more.

\begin{table*}[t]
\centering
\caption{\textbf{Ablation experiments.}
We perform experiments where we investigate the influence of different pre-training signals (a)-(c) and the addition of temporal differences at different points of the network (d)-(f). 
We report transfer learning performance to action recognition on UCF101 and HMDB51 and speed prediction accuracy on Kinetics-V.
}\label{tab:ablations}
\begin{tabular}{@{}l@{\hspace{.5em}}l@{\hspace{1em}}c@{\hspace{1em}}c@{\hspace{1em}}c@{\hspace{1em}}c@{\hspace{1em}}c@{}}
\toprule
 & & \multicolumn{2}{c}{\textbf{UCF101}} & \multicolumn{2}{c}{\textbf{HMDB51}} & \textbf{Kinetics-V} \\ 
\multicolumn{2}{l}{\textbf{Ablation}} & Linear & 1-NN & Linear  & 1-NN  & Speed Acc. \\ \midrule
(a) & uniform speed & 61.9 &  37.3 & 28.0 &  15.0 & 78.5 \\  
(b) & varying speed & 66.0 &  40.8 & 32.7 &  17.1 & 84.1 \\  
(c) & varying speed + SimCRL & 75.7 &  49.6 & 43.5 &  23.8  &  82.2 \\ 
\midrule
(d) & w/o temp-diff. & 66.0 &  40.8 & 32.7 &  17.1 & 84.1 \\  
(e) & input temp-diff. & 60.8 &  37.2 & 29.3 &  16.6 & 85.5 \\  
(f) & feature temp-diff. & 69.2 &  44.7 & 36.4 &  19.4  & 84.0  \\  

\bottomrule 
\end{tabular}
\label{tab.ablations}
\end{table*}

\subsection{Qualitative Evaluation via User Studies}\label{sec:user}

Finally, we performed a set of user studies comparing video re-timing based on our temporally varying speediness model to naive uniform video re-timing and re-timing based on frame similarity. 

\noindent \textbf{Study Setup.} To perform these studies, we re-timed 33 videos taken from the videos that make up the Vimeo-90K dataset \cite{xue2019video}.
We select videos with a length of around 3 minutes and ensure that the videos contain some dynamic object motion (typically humans moving or talking) and exclude videos that consist primarily of static landscape scenes, time-lapse footage, or similar to prevent ambiguous cases. 
To perform the study, we then randomly extract short ten-second clips from the longer re-timed videos and then ask users to compare two (approximately) corresponding clips for two re-timing techniques.
In total, we gather ten responses per clip, resulting in a sample size of 330. 
We report both the average response for all the possible answers and the preference after majority decision per clip. 
\\

\noindent \textbf{Naturalness.}
In the first study, we compare the results of re-timing with our temporally varying speediness model to a naive uniform speed-up of the video and ask participants to rate the naturalness of the videos. 
Given the two clips, we ask, "Which of the two videos looks more natural (i.e., has natural motion, is less hectic)?".
We also provide the option "Can not tell." as an answer.
The results of this study are summarized in Table~\ref{tab:study1}.
We observe that our re-timing approach is judged more natural for around 85\% of the video clips in the study.\\

\begin{table}[t]
    \centering
    \caption{\textbf{Naturalness user study.}
    We ask participants to compare video clips re-timed using our method to clips with naive uniform speed-up and indicate which one looks more natural, \ie, has more natural motion patterns. 
    We report both the average response as well as the majority vote per clip. 
    }
    \begin{tabular}{@{}l@{\hspace{2em}}c@{\hspace{1em}}c@{}}
    \toprule
    \textbf{Response}  &   \textbf{Average}  &   \textbf{Majority}  \\ 
    \midrule
    Uniform       & 27.0\%  & 15.2\%  \\
    Video-ReTime  & 64.5\%  & 84.8\% \\
    Can not tell  & \phantom{ }8.5\%  &  \phantom{ }0.0\% \\
    \bottomrule
    \end{tabular}
    \label{tab:study1}
\end{table}

\noindent \textbf{Re-Timing Signal Preference.}
In the second study, we compare re-timing based on the speediness model to a re-timing based on frame feature similarity. 
Given the two re-timed clips, we ask, "Which of the two video edits do you prefer?".
We again provide the option "Can not tell." as an answer.
The results of this study are summarized in Table~\ref{tab:study2}.
While speediness-based re-timing is preferred in the majority of the videos (62\%), some users do tend to prefer re-timing based on frame similarity. 
Providing alternative re-timing signals that match the user's subjective preferences through the formulation in Section \ref{sec:general} could thus be very useful.

\begin{table}[t]
    \centering
    \caption{\textbf{Re-timing signal preference study.} 
    We ask participants to indicate their preference when comparing video clips re-timed based on speediness predictions and clips re-timed based on consecutive frame feature similarity.
    We report both the average response as well as the majority vote per clip. 
    }
    \begin{tabular}{@{}l@{\hspace{2em}}c@{\hspace{1em}}c@{}}
    \toprule
    \textbf{Response}  &   \textbf{Average}  &   \textbf{Majority}  \\ 
    \midrule
    Feature similarity       & 43.2\%  & 37.9\%  \\
    Speediness  & 52.4\%  &  62.1\% \\
    No preference  & \phantom{ }4.4\%  &  \phantom{ }0.0\% \\
    \bottomrule
    \end{tabular}
    \label{tab:study2}
\end{table}

\begin{figure*}
    \centering
    \includegraphics[width=0.85\linewidth]{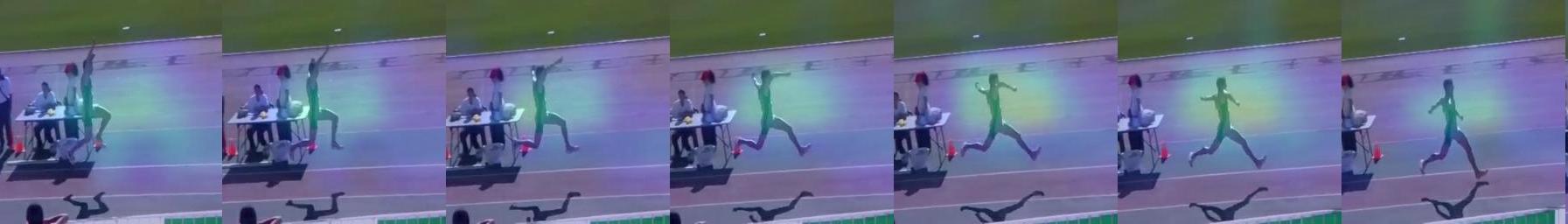}
    \includegraphics[width=0.85\linewidth]{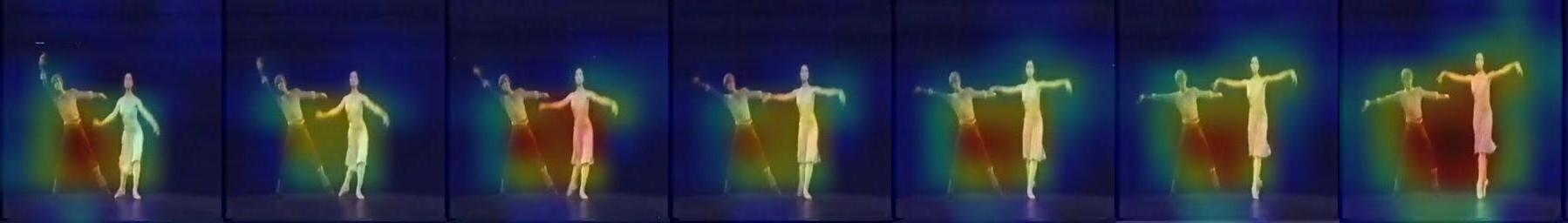}
    \includegraphics[width=0.85\linewidth]{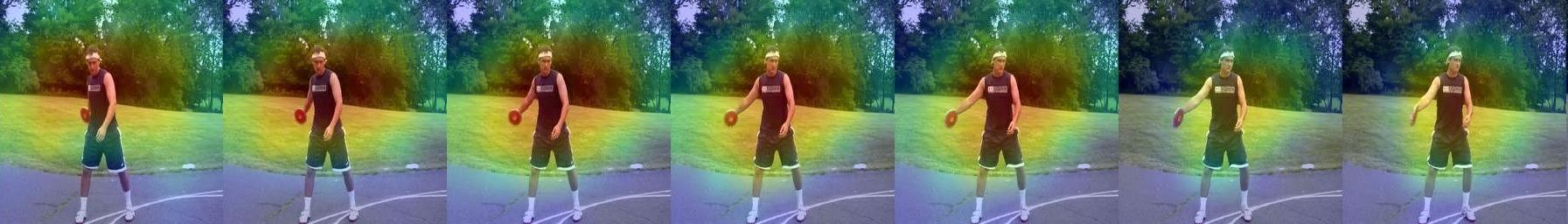}
    \includegraphics[width=0.85\linewidth]{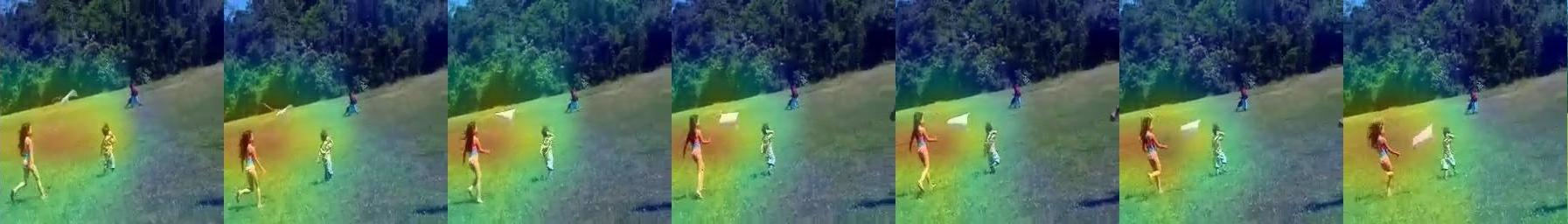}
    \includegraphics[width=0.85\linewidth]{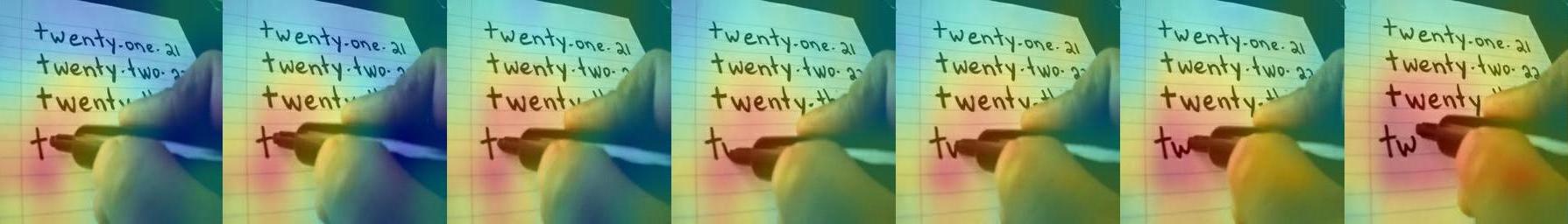}

    \caption{\textbf{Grad-CAM visualization of the speediness predictions.}
    We use Grad-CAM to obtain visual explanations for the speediness predictions of our model.
    We can observe strong responses in spatio-temporal regions with foreground object movement and a strong focus on human body pose.  
    This suggests that the model primarily focuses on the motion patterns of objects rather than low-level or background cues. 
    }
    \label{fig:gradcam}
\end{figure*}

\begin{figure*}
    \centering
    \includegraphics[width=0.28\linewidth]{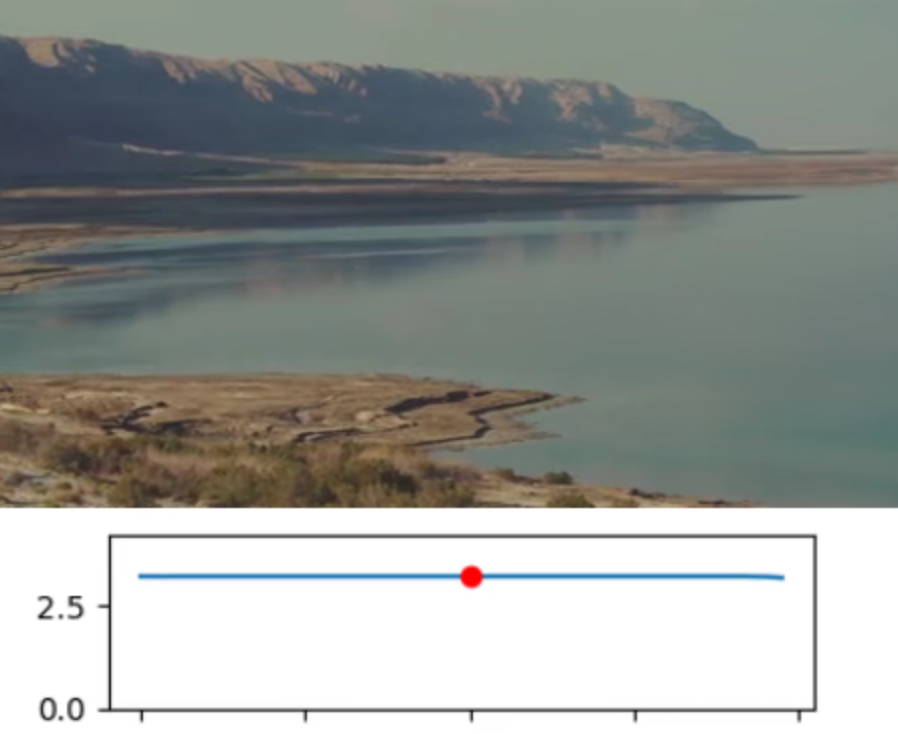}
    \includegraphics[width=0.28\linewidth]{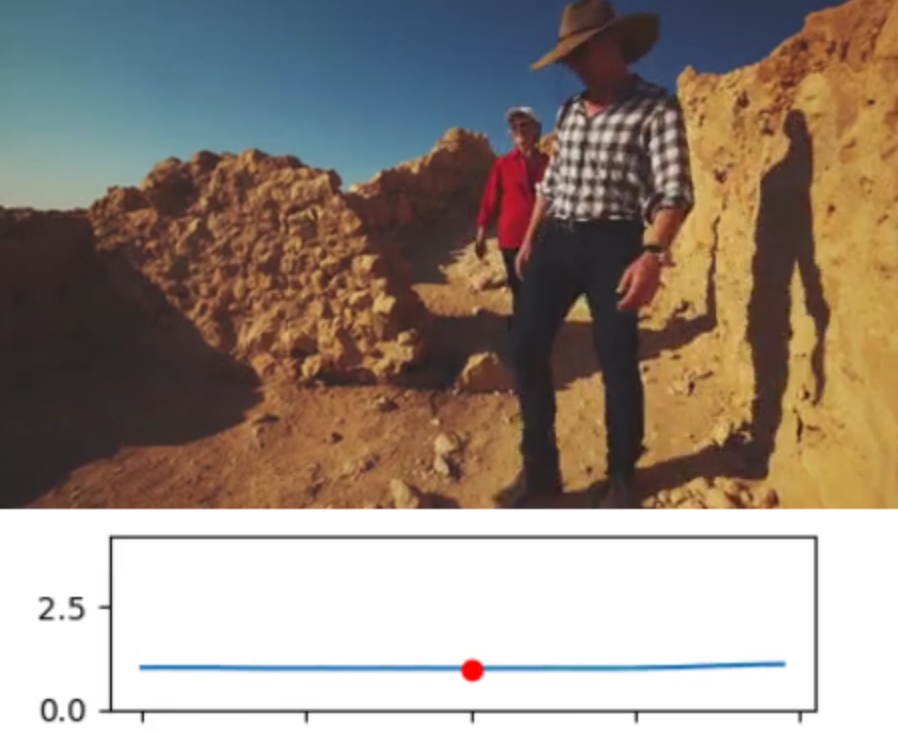}
    \includegraphics[width=0.28\linewidth]{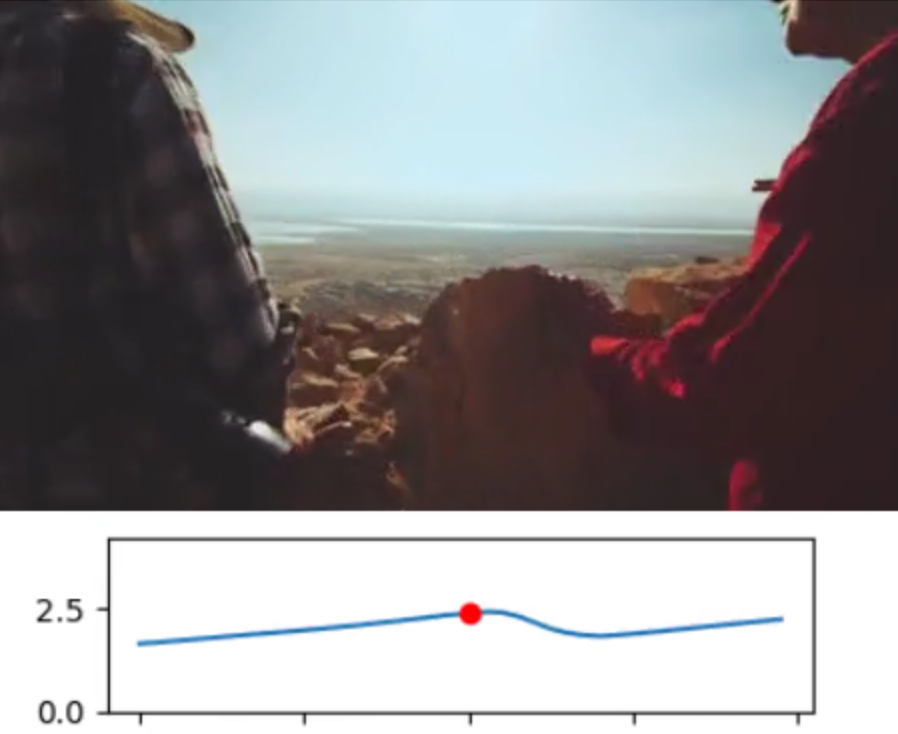}
    \caption{\textbf{Illustration of re-timed scenes.}
    We show example frames and per-frame speed-up curves (below) for one of the videos in our user study. 
    The scene on the left shows camera-panning, which is speed-up.
    In the middle is a scene with people moving, which is kept at the original natural speed.
    We again see camera motion on the right, this time sweeping in between the persons standing still. 
    }
    \label{fig:curve}
\end{figure*}

\subsection{Visualization}
Finally, to better understand what the speediness model focuses on when it decides the speed of frames, we visualize the most important spatio-temporal regions in the input frames using Grad-CAM \cite{selvaraju2017grad}.
The results on some examples of Kinetics-U can be found in Figure~\ref{fig:gradcam}. 
We can primarily observe a strong focus of the model on the foreground object. 
We believe this is a good sign, as it suggests that the model relies less on background or low-level cues (\eg, camera shake) and instead learns the natural movement patterns of the involved objects. 

We also show example scenes and adaptive speed-up curves for one of the videos of our user study in Figure~\ref{fig:curve}.
We notice that the method speeds up scenes without object movement even when they contain large optical flow (\eg, due to moving cameras) while preserving scenes where people are moving or speaking at their normal speed. 


\section{Conclusions}

We introduced a method to learn time-varying frame speediness through self-supervision for time re-mapping.
Our model learns a probability distribution over different per frame playback speeds given a sequence of input frames.
We then leverage this speed model to define a re-timing signal which can be used to reduce the duration of videos while maximally preserving the natural motion patterns in the video. 
To this end, we formulate an optimization problem that optimizes the frame subsampling to match a user-specified target duration and matches the re-timing signal (\eg, based on per-frame speediness), which constrains the relative speedup applied to different frames of the video.
Experiments demonstrate that our model and formulation is more accurate at localizing changes in the playback speed, performs better on longer videos, and has a lower run-time than prior work.

{\small
\bibliographystyle{ieee_fullname}
\bibliography{egbib}

\begin{thebibliography}{10}\itemsep=-1pt

\bibitem{apostolidis2021video}
Evlampios Apostolidis, Eleni Adamantidou, Alexandros~I Metsai, Vasileios
  Mezaris, and Ioannis Patras.
\newblock Video summarization using deep neural networks: A survey.
\newblock {\em Proceedings of the IEEE}, 109(11):1838--1863, 2021.

\bibitem{bai2020can}
Yutong Bai, Haoqi Fan, Ishan Misra, Ganesh Venkatesh, Yongyi Lu, Yuyin Zhou,
  Qihang Yu, Vikas Chandra, and Alan Yuille.
\newblock Can temporal information help with contrastive self-supervised
  learning?
\newblock {\em arXiv preprint arXiv:2011.13046}, 2020.

\bibitem{benaim2020speednet}
Sagie Benaim, Ariel Ephrat, Oran Lang, Inbar Mosseri, William~T Freeman,
  Michael Rubinstein, Michal Irani, and Tali Dekel.
\newblock Speednet: Learning the speediness in videos.
\newblock In {\em Proceedings of the IEEE/CVF Conference on Computer Vision and
  Pattern Recognition}, pages 9922--9931, 2020.

\bibitem{bennett2007computational}
Eric~P Bennett and Leonard McMillan.
\newblock Computational time-lapse video.
\newblock In {\em ACM SIGGRAPH 2007 papers}, pages 102--es. 2007.

\bibitem{brattoli2017lstm}
Biagio Brattoli, Uta B{\"u}chler, Anna-Sophia Wahl, Martin~E Schwab, and
  Bj{\"o}rn Ommer.
\newblock Lstm self-supervision for detailed behavior analysis.
\newblock In {\em Proceedings of the IEEE Conference on Computer Vision and
  Pattern Recognition (CVPR)}, volume~2, 2017.

\bibitem{chen2020simple}
Ting Chen, Simon Kornblith, Mohammad Norouzi, and Geoffrey Hinton.
\newblock A simple framework for contrastive learning of visual
  representations.
\newblock In {\em International conference on machine learning}, pages
  1597--1607. PMLR, 2020.

\bibitem{dave2021tclr}
Ishan Dave, Rohit Gupta, Mamshad~Nayeem Rizve, and Mubarak Shah.
\newblock Tclr: Temporal contrastive learning for video representation.
\newblock {\em arXiv preprint arXiv:2101.07974}, 2021.

\bibitem{Carl2015}
Carl Doersch, Abhinav Gupta, and Alexei~A. Efros.
\newblock Unsupervised visual representation learning by context prediction.
\newblock {\em ICCV}, 2015.

\bibitem{epstein2020oops}
Dave Epstein, Boyuan Chen, and Carl Vondrick.
\newblock Oops! predicting unintentional action in video.
\newblock In {\em Proceedings of the IEEE/CVF Conference on Computer Vision and
  Pattern Recognition}, pages 919--929, 2020.

\bibitem{fernando2017self}
Basura Fernando, Hakan Bilen, Efstratios Gavves, and Stephen Gould.
\newblock Self-supervised video representation learning with odd-one-out
  networks.
\newblock In {\em Computer Vision and Pattern Recognition (CVPR), 2017 IEEE
  Conference on}, pages 5729--5738. IEEE, 2017.

\bibitem{gidaris2018unsupervised}
Spyros Gidaris, Praveer Singh, and Nikos Komodakis.
\newblock Unsupervised representation learning by predicting image rotations.
\newblock In {\em International Conference on Learning Representations}, 2018.

\bibitem{he2020momentum}
Kaiming He, Haoqi Fan, Yuxin Wu, Saining Xie, and Ross Girshick.
\newblock Momentum contrast for unsupervised visual representation learning.
\newblock In {\em Proceedings of the IEEE/CVF Conference on Computer Vision and
  Pattern Recognition}, pages 9729--9738, 2020.

\bibitem{jenni2021time}
Simon Jenni and Hailin Jin.
\newblock Time-equivariant contrastive video representation learning.
\newblock In {\em Proceedings of the IEEE/CVF International Conference on
  Computer Vision}, pages 9970--9980, 2021.

\bibitem{jenni2020steering}
Simon Jenni, Hailin Jin, and Paolo Favaro.
\newblock Steering self-supervised feature learning beyond local pixel
  statistics.
\newblock In {\em Proceedings of the IEEE/CVF Conference on Computer Vision and
  Pattern Recognition}, pages 6408--6417, 2020.

\bibitem{jenni2020video}
Simon Jenni, Givi Meishvili, and Paolo Favaro.
\newblock Video representation learning by recognizing temporal
  transformations.
\newblock {\em arXiv preprint arXiv:2007.10730}, 2020.

\bibitem{hmdb51}
H. Kuehne, H. Jhuang, E. Garrote, T. Poggio, and T. Serre.
\newblock {HMDB}: a large video database for human motion recognition.
\newblock In {\em Proceedings of the International Conference on Computer
  Vision (ICCV)}, 2011.

\bibitem{lee2017unsupervised}
Hsin-Ying Lee, Jia-Bin Huang, Maneesh Singh, and Ming-Hsuan Yang.
\newblock Unsupervised representation learning by sorting sequences.
\newblock In {\em Proceedings of the IEEE International Conference on Computer
  Vision}, pages 667--676, 2017.

\bibitem{loshchilov2016sgdr}
Ilya Loshchilov and Frank Hutter.
\newblock Sgdr: Stochastic gradient descent with warm restarts.
\newblock {\em arXiv preprint arXiv:1608.03983}, 2016.

\bibitem{loshchilov2018decoupled}
Ilya Loshchilov and Frank Hutter.
\newblock Fixing weight decay regularization in adam.
\newblock {\em arXiv preprint arXiv:1711.05101}, 2017.

\bibitem{misra2016shuffle}
Ishan Misra, C~Lawrence Zitnick, and Martial Hebert.
\newblock Shuffle and learn: unsupervised learning using temporal order
  verification.
\newblock In {\em European Conference on Computer Vision}, pages 527--544.
  Springer, 2016.

\bibitem{noroozi2016unsupervised}
Mehdi Noroozi and Paolo Favaro.
\newblock Unsupervised learning of visual representations by solving jigsaw
  puzzles.
\newblock In {\em European Conference on Computer Vision}, pages 69--84.
  Springer, 2016.

\bibitem{patrick2020multi}
Mandela Patrick, Yuki~M Asano, Ruth Fong, Jo{\~a}o~F Henriques, Geoffrey Zweig,
  and Andrea Vedaldi.
\newblock Multi-modal self-supervision from generalized data transformations.
\newblock {\em arXiv preprint arXiv:2003.04298}, 2020.

\bibitem{petrovic2005adaptive}
Nemanja Petrovic, Nebojsa Jojic, and Thomas~S Huang.
\newblock Adaptive video fast forward.
\newblock {\em Multimedia Tools and Applications}, 26(3):327--344, 2005.

\bibitem{poleg2015egosampling}
Yair Poleg, Tavi Halperin, Chetan Arora, and Shmuel Peleg.
\newblock Egosampling: Fast-forward and stereo for egocentric videos.
\newblock In {\em Proceedings of the IEEE Conference on Computer Vision and
  Pattern Recognition}, pages 4768--4776, 2015.

\bibitem{selvaraju2017grad}
Ramprasaath~R Selvaraju, Michael Cogswell, Abhishek Das, Ramakrishna Vedantam,
  Devi Parikh, and Dhruv Batra.
\newblock Grad-cam: Visual explanations from deep networks via gradient-based
  localization.
\newblock In {\em Proceedings of the IEEE international conference on computer
  vision}, pages 618--626, 2017.

\bibitem{smith1997video}
Michael~A Smith and Takeo Kanade.
\newblock Video skimming and characterization through the combination of image
  and language understanding techniques.
\newblock In {\em Proceedings of IEEE Computer Society Conference on Computer
  Vision and Pattern Recognition}, pages 775--781. IEEE, 1997.

\bibitem{soomro2012ucf101}
Khurram Soomro, Amir~Roshan Zamir, and Mubarak Shah.
\newblock Ucf101: A dataset of 101 human actions classes from videos in the
  wild.
\newblock {\em arXiv preprint arXiv:1212.0402}, 2012.

\bibitem{tran2018closer}
Du Tran, Heng Wang, Lorenzo Torresani, Jamie Ray, Yann LeCun, and Manohar
  Paluri.
\newblock A closer look at spatiotemporal convolutions for action recognition.
\newblock In {\em Proceedings of the IEEE conference on Computer Vision and
  Pattern Recognition}, pages 6450--6459, 2018.

\bibitem{wei2018learning}
Donglai Wei, Joseph Lim, Andrew Zisserman, and William~T Freeman.
\newblock Learning and using the arrow of time.
\newblock In {\em Proceedings of the IEEE Conference on Computer Vision and
  Pattern Recognition}, pages 8052--8060, 2018.

\bibitem{xue2019video}
Tianfan Xue, Baian Chen, Jiajun Wu, Donglai Wei, and William~T Freeman.
\newblock Video enhancement with task-oriented flow.
\newblock {\em International Journal of Computer Vision (IJCV)},
  127(8):1106--1125, 2019.

\bibitem{yao2020video}
Yuan Yao, Chang Liu, Dezhao Luo, Yu Zhou, and Qixiang Ye.
\newblock Video playback rate perception for self-supervised spatio-temporal
  representation learning.
\newblock In {\em Proceedings of the IEEE/CVF Conference on Computer Vision and
  Pattern Recognition}, pages 6548--6557, 2020.

\bibitem{zhang2016colorful}
Richard Zhang, Phillip Isola, and Alexei~A Efros.
\newblock Colorful image colorization.
\newblock In {\em European Conference on Computer Vision}, pages 649--666.
  Springer, 2016.

\bibitem{zhang2016split}
Richard Zhang, Phillip Isola, and Alexei~A Efros.
\newblock Split-brain autoencoders: Unsupervised learning by cross-channel
  prediction.
\newblock In {\em Proceedings of the IEEE Conference on Computer Vision and
  Pattern Recognition}, pages 1058--1067, 2017.

\bibitem{zhou2014time}
Feng Zhou, Sing Bing~Kang, and Michael~F Cohen.
\newblock Time-mapping using space-time saliency.
\newblock In {\em proceedings of the IEEE Conference on Computer Vision and
  Pattern Recognition}, pages 3358--3365, 2014.

\bibitem{zisserman2017kinetics}
Andrew Zisserman, Joao Carreira, Karen Simonyan, Will Kay, Brian Zhang, Chloe
  Hillier, Sudheendra Vijayanarasimhan, Fabio Viola, Tim Green, Trevor Back,
  et~al.
\newblock The kinetics human action video dataset.
\newblock {\em ArXiv}, 2017.

\end{thebibliography}
}

\end{document}